\documentclass[a4paper,11pt]{article}

\usepackage[utf8]{inputenc}
\usepackage[T1]{fontenc}
\usepackage{amsmath,amssymb,amsfonts}
\usepackage{algorithmic}
\usepackage{graphicx,color}
\usepackage{textcomp}
\usepackage{booktabs}
\usepackage{multirow}
\usepackage{caption}
\usepackage{subcaption}
\usepackage{algorithm}
\usepackage{array}
\usepackage[margin=1in]{geometry} 
\usepackage{url}
\usepackage{verbatim}
\usepackage{hyperref}             
\usepackage{enumitem}             
\usepackage{fancyhdr}             

\title{\bfseries STAGformer: A Spatio-temporal Agent Graph Transformer\\ for Micro Mobility Demand Forecasting}

\author{
    Ye Zihao\\[4pt]
    \small Department of System Engineering\\City University of Hong Kong\\[2pt]
}
\date{}

\begin{document}

\maketitle

\begin{abstract}
\noindent
Accurate station-level demand forecasting is essential for the efficient operation of bike-sharing systems, yet it remains challenging due to complex spatio-temporal dependencies and the large scale of urban networks. This paper presents STAGformer, a Spatio-Temporal Agent Graph Transformer that achieves efficient global modeling with linear computational complexity. The model introduces a two-step agent attention mechanism, where a small set of learnable spatial and temporal agent tokens first aggregate global information and then broadcast it back to individual stations and time steps, effectively capturing long-range interactions while reducing the quadratic cost of standard self-attention to \(\mathcal{O}(NT)\). STAGformer integrates four core modules: a spatio-temporal encoder that fuses dynamic node features with external contextual factors (weather, time, points of interest), a graph propagation module for spatial neighbor aggregation, a temporal convolution module for local pattern extraction, and the agent attention module for global dependency modeling. Extensive experiments on two real-world datasets---NYC Citi-Bike and Chicago Divvy-Bike---demonstrate that STAGformer consistently outperforms state-of-the-art baselines across multiple prediction horizons, achieving significant improvements in both RMSE and MAE. Ablation studies validate the contribution of each component, with the agent attention mechanism proving critical for modeling global spatio-temporal dependencies. By combining the expressive power of softmax attention with linear efficiency, STAGformer provides a scalable and accurate solution for micro-mobility demand forecasting in large-scale urban systems.
\end{abstract}

\vspace{4pt}
\noindent\textbf{Keywords:} Bike-sharing, Graph Neural Network, Traffic forecasting, Demand modelling

\vspace{12pt}

\section{Introduction}

\begin{figure}[!htbp]
    \centering
    \includegraphics[width=0.85\textwidth]{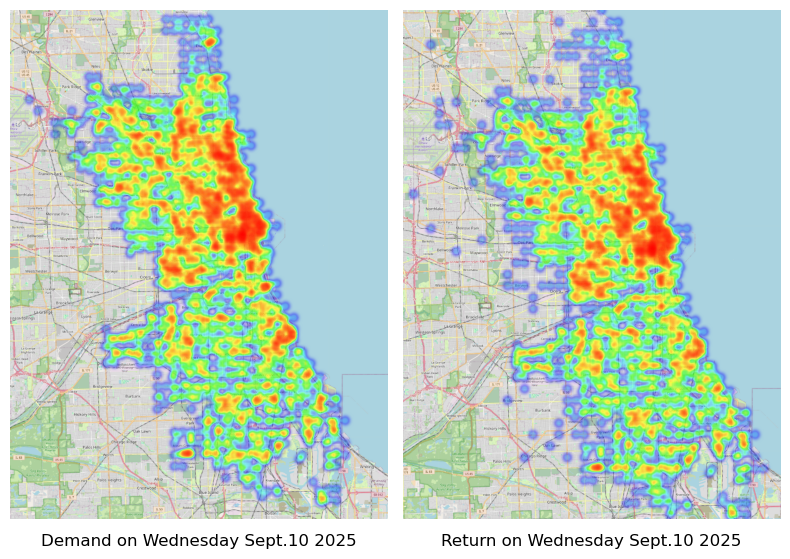}
    \caption{Spatial imbalance of bike-sharing demand and returns in Chicago on Wednesday, September 10, 2025. Red areas indicate high demand or returns, blue areas low activity.}
    \label{fig:chicago}
\end{figure}

As a green and flexible mode of micro-mobility, bike-sharing has been widely adopted in numerous cities worldwide, effectively alleviating traffic congestion and promoting low-carbon travel \cite{faghih2016incorporating, behroozi2025predicting}. However, bike-sharing systems are inherently dynamic and spatio-temporally heterogeneous: rental and return demands at stations exhibit periodic fluctuations over time, while spatially they are influenced by road network structures, the distribution of points of interest (POIs), and real-time weather conditions (as illustrated in Fig.~\ref{fig:chicago}). Therefore, accurate station-level demand forecasting is crucial for vehicle rebalancing, operational optimization, and enhancing user experience \cite{wang2023demand, feng2024adaptive}.

In recent years, deep learning techniques, particularly Graph Neural Networks (GNNs) and Transformer architectures, have significantly advanced spatio-temporal forecasting \cite{li2017diffusion, yu2017spatio, vaswani2017attention}. GNNs effectively model spatial dependencies among stations \cite{wu2019graph, bai2020adaptive}, while Transformers excel at capturing long-range temporal correlations \cite{xu2020spatial, zhou2021informer}. Nevertheless, existing methods face two major challenges when applied to large-scale bike-sharing networks: (i) the computational complexity of standard self-attention scales quadratically with the product of the number of stations \(N\) and time steps \(T\), i.e., \(\mathcal{O}((NT)^2)\), rendering it impractical for city-scale systems \cite{vaswani2017attention, child2019generating}; (ii) most models fail to effectively integrate multi-source external factors (e.g., weather, POIs) that influence demand, leading to limited prediction accuracy \cite{rochas2023contextual, xiang2025bike}.

To address these challenges, we propose a Spatio-Temporal Agent Graph Transformer (STAGformer) based on an agent attention mechanism. By introducing a small set of learnable ``agent'' tokens, STAGformer reduces the complexity of self-attention to linear in both the number of stations and time steps, thereby efficiently capturing global spatio-temporal dependencies \cite{han2024agent}. Specifically, STAGformer comprises four core modules: a spatio-temporal feature encoder that projects historical station features and external factors into a high-dimensional representation and injects positional information; a graph propagation module that explicitly aggregates information from spatial neighbors using the graph structure; a temporal convolution module that extracts local temporal patterns for each station; and an agent attention module that models long-range interactions between stations and time steps with linear complexity through a two-step aggregation and broadcasting mechanism. Finally, the fused features are passed through an output layer to generate multi-step predictions of rental and return demands.

The main contributions of this paper are as follows:
\begin{enumerate}[leftmargin=*]
    \item We design an encoding scheme that fuses dynamic node features with global external features, and enhance both global and local pattern extraction through graph propagation and temporal convolution.
    \item We propose STAGformer, which introduces the agent attention mechanism to the task of bike-sharing demand forecasting, achieving linear computational complexity while maintaining global modeling capability.
    \item Extensive experiments on two real-world datasets demonstrate that STAGformer outperforms state-of-the-art baselines in terms of RMSE and MAE, while offering higher computational efficiency.
\end{enumerate}

The remainder of this paper is organized as follows: Section~2 reviews related work; Section~3 formulates the problem and provides preliminaries; Section~4 details the STAGformer architecture; Section~5 describes the experimental setup and results analysis; Section~6 concludes the paper and discusses future work.

\section{Related Work}

\subsection{Spatio-Temporal Forecasting: Evolution and Challenges}

Station-level demand forecasting in micro-mobility systems, such as bike-sharing, is inherently challenging due to complex spatial dependencies, temporal dynamics, and the large scale of urban networks. Early approaches primarily relied on time series models, e.g., ARIMA and its seasonal variants SARIMA, which capture periodicities and trends but fail to incorporate spatial correlations and external factors \cite{chatfield2019analysis, kim2025comprehensive}. Subsequently, machine learning models like Support Vector Regression (SVR), Random Forests, and Gradient Boosting Trees were adopted, improving accuracy by manually engineering features (e.g., weather, holidays) \cite{friedman2001greedy, zhou2021machine}. However, these methods cannot automatically mine spatial dependencies among stations and have limited capacity for long-range temporal modeling.

The advent of deep learning revolutionized spatio-temporal forecasting. Recurrent Neural Networks (RNNs) and their variants (LSTM, GRU) were widely applied to model time series for each station independently, effectively capturing non-linear temporal dependencies \cite{hochreiter1997long, cho2014learning}. To simultaneously model spatial correlations, researchers turned to Convolutional Neural Networks (CNNs) applied to grid-based urban representations \cite{zhang2017deep}, but grid partitioning fails to adapt to non-Euclidean road network structures. The emergence of Graph Neural Networks (GNNs) addressed this issue. The Diffusion Convolutional Recurrent Network (DCRNN) \cite{li2017diffusion} combined diffusion convolution with a sequence-to-sequence architecture, achieving the first spatio-temporal prediction on graphs. The Spatio-Temporal Graph Convolutional Network (STGCN) \cite{yu2017spatio} utilized alternating stacks of 1D temporal convolutions and graph convolutions, significantly improving computational efficiency.

Subsequent studies further explored dynamic graph modeling: ASTGCN introduced attention mechanisms to adaptively adjust spatial weights \cite{guo2019attention}; GMAN designed an encoder-decoder structure with spatio-temporal attention modules to handle long-range dependencies \cite{zheng2020gman}. For large-scale dynamic graphs, TrafficStream proposed a streaming GNN framework with continual learning \cite{chen2021trafficstream}. More recent works such as Graph WaveNet \cite{wu2019graph} and AGCRN \cite{bai2020adaptive} incorporate adaptive graph learning to capture hidden spatial dependencies without relying on a predefined graph structure. These methods excel in modeling local spatio-temporal dependencies but are often limited by fixed or static graph structures and still struggle with capturing global long-range interactions efficiently.

\subsection{Attention Mechanisms and Transformer-based Models}

The attention mechanism, first achieving breakthroughs in natural language processing \cite{vaswani2017attention}, was soon introduced to spatio-temporal forecasting tasks. Transformers, with their self-attention capability to directly model dependencies between any two positions, have shown strong potential in traffic flow prediction \cite{xu2020spatial}. However, the standard self-attention mechanism incurs a computational complexity of \(\mathcal{O}((NT)^2)\), where \(N\) is the number of nodes and \(T\) is the number of time steps, making it prohibitive for city-scale networks.

To mitigate this, several efficient attention variants have been proposed:
\begin{itemize}[leftmargin=*]
    \item \textbf{Sparse attention} restricts each query to attend to only a subset of keys, reducing complexity but potentially sacrificing global context \cite{child2019generating}.
    \item \textbf{Separated attention} performs spatial and temporal attention separately, e.g., first attending among nodes per time step (\(\mathcal{O}(TN^2)\)) and then along time per node (\(\mathcal{O}(NT^2)\)). While reducing cost compared to joint attention, it still involves quadratic terms in \(N\) or \(T\), which remains problematic when either dimension is large \cite{zheng2020gman, guo2019attention}.
    \item \textbf{Linear attention} replaces the Softmax similarity with kernel functions to achieve linear complexity, but often suffers from degraded expressiveness and difficulty in capturing sharp attention patterns \cite{katharopoulos2020transformers, choromanski2020rethinking}.
\end{itemize}

\begin{figure}[!t]
    \centering
    \includegraphics[width=\textwidth]{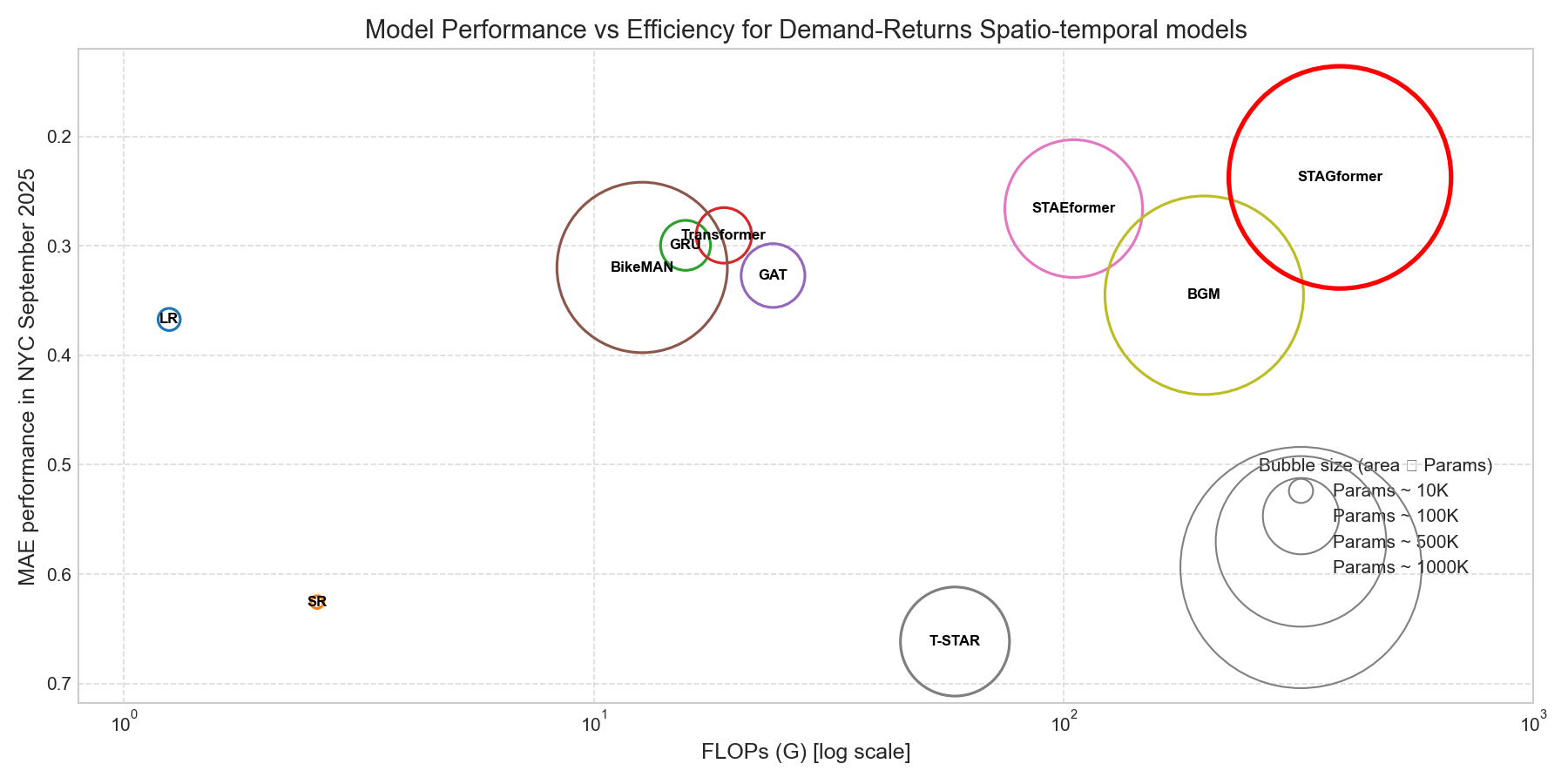}
    \caption{Comparison of model performance and efficiency (NYC, September 2025, MAE; bubble size indicates parameter count).}
    \label{fig:model_comparison}
\end{figure}

In parallel, Graph Attention Networks (GATs) have achieved remarkable success on graph-structured data by assigning different weights to neighboring nodes via attention \cite{velivckovic2017graph, brody2021attentive, zhang2018gaan}. In spatio-temporal forecasting, many studies combine GAT with recurrent or convolutional networks; ASTGCN \cite{guo2019attention} and GMAN \cite{zheng2020gman} utilize attention to enhance spatial dependency modeling, while MTGNN \cite{wu2020connecting} captures multi-scale spatio-temporal correlations through graph attention layers. These methods improve modeling of dynamic spatial relationships but still suffer from high computational complexity, limiting their scalability to large station networks.

More recently, Transformer variants designed for long sequences, such as Informer \cite{zhou2021informer}, have been applied to time series forecasting. However, they primarily focus on temporal dimension efficiency and do not address the joint spatio-temporal scalability challenge. Set Transformers \cite{lee2019set} introduce inducing points to reduce complexity in set-structured data, which shares conceptual similarity with our agent-based approach, but they are not tailored to spatio-temporal graphs and do not incorporate explicit spatial graph propagation.

\subsection{Design Rationale and Advantages of STAGformer}

Despite the significant progress made by the aforementioned methods, existing models still struggle to balance computational efficiency and global modeling capability for large-scale bike-sharing networks. STAGformer addresses these gaps by leveraging the agent attention mechanism~\cite{han2024agent} in the spatio-temporal domain and integrating it with complementary modules. The key design features of our approach are:

\begin{itemize}[leftmargin=*]
    \item \textbf{Linear complexity with global receptive field:} The agent attention mechanism~\cite{han2024agent} replaces the quadratic self-attention with a two-step aggregation-broadcasting process mediated by a small set of learnable agent tokens, reducing complexity from \(\mathcal{O}((NT)^2)\) to \(\mathcal{O}(NT)\). We extend this mechanism to spatio-temporal data by introducing separate spatial and temporal agent tokens that independently capture cross-station and cross-time interactions, while preserving the Softmax attention's ability to model arbitrary long-range dependencies---unlike sparse or kernel-based linear attention methods that may compromise global information exchange.
    
    \item \textbf{Preservation of expressiveness:} Unlike linear attention variants that replace Softmax with kernel functions and often sacrifice representational capacity, STAGformer inherits the full Softmax attention in both agent-attention steps and employs a depthwise convolution (DWC) residual branch~\cite{han2024agent} to compensate for potential feature diversity loss. This design ensures that the efficiency gains do not come at the cost of predictive accuracy.
    
    \item \textbf{Integration of local and global modules:} While agent attention captures global dependencies, STAGformer also incorporates a graph propagation module for explicit spatial neighbor aggregation and a temporal convolution module for local temporal pattern extraction. This hybrid design complements the global modeling with fine-grained local information, which is often missing in pure Transformer architectures.
    
    \item \textbf{Multi-source feature fusion:} STAGformer seamlessly integrates dynamic node features with external factors (weather, time, POIs) through a dedicated encoding scheme, enhancing its ability to capture contextual influences ignored by many existing models.
\end{itemize}

In summary, STAGformer provides a scalable and accurate solution for micro-mobility demand forecasting by combining the strengths of graph-based local modeling and attention-based global modeling, all with linear computational complexity. The experimental results on real-world datasets demonstrate its superiority over a wide range of baseline methods, confirming the effectiveness of the proposed approach.

\section{Preliminary}

\subsection{Bike-sharing System}

An urban bike-sharing network can be modeled as a graph \(\mathcal{G} = (\mathcal{V}, \mathcal{E})\), where \(\mathcal{V} = \{v_1, v_2, \dots, v_N\}\) denotes the set of \(N\) rental stations and \(\mathcal{E}\) represents the spatial connections between stations. At each time step \(t\), every station \(v_i \in \mathcal{V}\) records multi-dimensional status information, such as the number of rentals, returns, and available bikes, forming a node feature vector \(\mathbf{x}_{i,t} \in \mathbb{R}^{F_n}\). Aggregating features of all stations at time \(t\) yields a matrix \(\mathbf{X}_t \in \mathbb{R}^{N \times F_n}\), and the complete historical observation sequence can be represented as a 4D tensor \(\mathcal{X} \in \mathbb{R}^{B \times T \times N \times F_n}\), where \(B\) is the batch size and \(T\) is the length of the historical time window.

In addition to station status, bike-sharing demand is influenced by various external factors, such as weather conditions (temperature, precipitation, wind speed) and temporal information (hour of the day, day of the week, holidays). Let \(\mathbf{g}_t \in \mathbb{R}^{F_g}\) denote the global feature vector at time step \(t\); then the sequence of external factors can be expressed as \(\mathcal{G} \in \mathbb{R}^{B \times T \times F_g}\).

\subsection{Problem Definition}

We adopt a sliding window prediction paradigm. Given a historical observation window of length \(T\) and a prediction horizon \(H\), the goal is to learn a mapping function \(f\) that projects historical observations onto future station-level demands over \(H\) time steps. For an arbitrary batch, the problem is formalized as:

\begin{equation}
\hat{\mathcal{Y}}_{t+1:t+H} = f\left( \mathcal{X}_{t-T+1:t}, \mathcal{G}_{t-T+1:t}, \mathbf{A}_{\text{ext}} \right),
\label{eq:problem}
\end{equation}

where:
\(\hat{\mathcal{Y}}_{t+1:t+H} \in \mathbb{R}^{B \times H \times N \times D}\) denotes the predicted \(D\)-dimensional targets (e.g., rental and return counts) for each station over the future \(H\) steps;
\(\mathcal{X}_{t-T+1:t} \in \mathbb{R}^{B \times T \times N \times F_n}\) is the historical node features within the observation window;
\(\mathcal{G}_{t-T+1:t} \in \mathbb{R}^{B \times T \times F_g}\) is the global features (including weather and temporal information) within the observation window;
\(\mathbf{A}_{\text{ext}} \in \mathbb{R}^{N \times N}\) is an external adjacency matrix (e.g., road network distance matrix).

\section{Methodology}

\begin{figure}[!t]
    \centering
    \includegraphics[width=\textwidth]{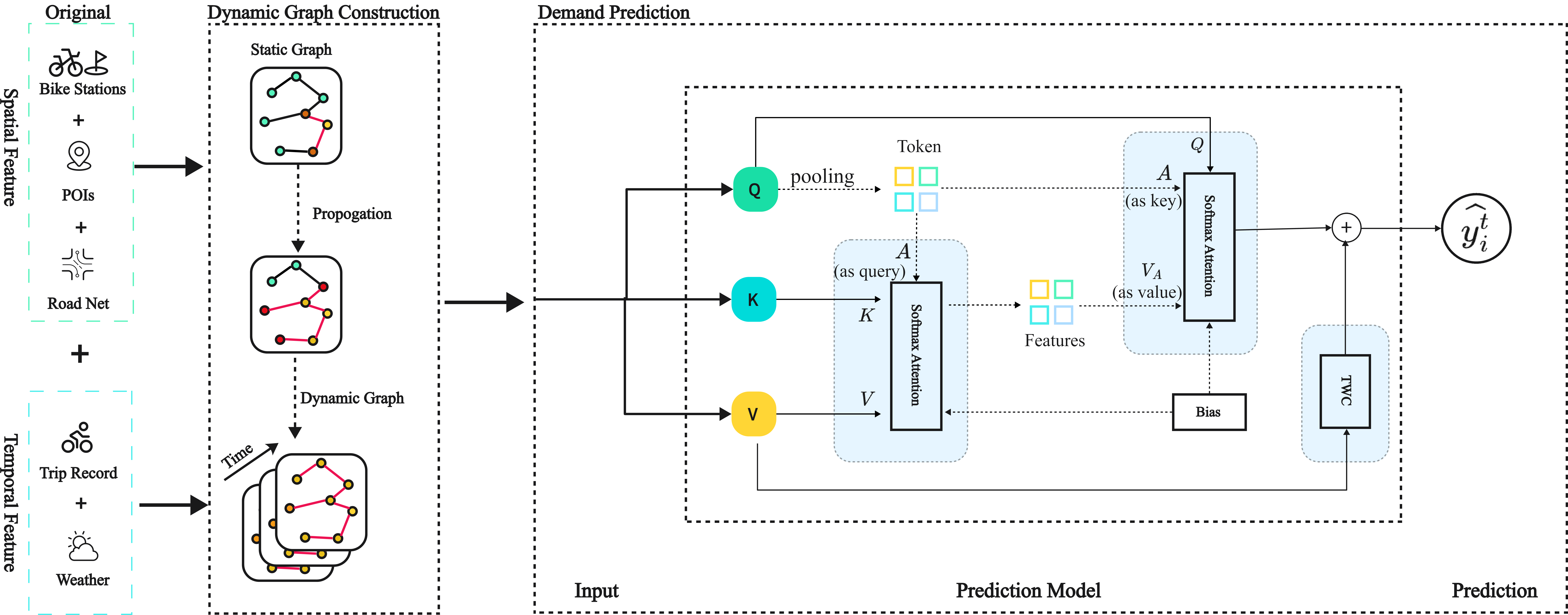}
    \caption{Overall architecture of \underline{S}patio-\underline{T}emporal \underline{A}gent \underline{G}raph Trans\underline{former} (STAGformer).}
    \label{fig:architecture}
\end{figure}

To improve the accuracy and robustness of station-level bike-sharing demand forecasting, we propose a Spatio-Temporal Agent Graph Transformer (STAGformer) based on agent attention. The model constructs multi-level feature extraction modules to capture temporal dependencies, spatial correlations, and global spatio-temporal interactions, while reducing the quadratic complexity of standard self-attention to linear through the agent attention mechanism. The overall framework, illustrated in Fig.~\ref{fig:architecture}, comprises four core modules: a spatio-temporal feature encoder that projects raw inputs into high-dimensional representations and injects positional information; a graph propagation module that explicitly aggregates influences from spatial neighbors using the graph structure; a temporal convolution module that models local temporal dependencies; and an agent attention module that efficiently captures global spatio-temporal correlations with linear complexity. Finally, the fused features are passed through an output layer to generate future station-level demand predictions.

\subsection{Spatio-Temporal Feature Encoding}

The model input consists of two types of sequential data: dynamic node features \(\mathcal{X} \in \mathbb{R}^{B \times T \times N \times F_n}\) and global features \(\mathcal{G} \in \mathbb{R}^{B \times T \times F_g}\). The feature encoding process is as follows:

\begin{enumerate}[leftmargin=*]
    \item \textbf{Linear projection}: Apply linear transformations to node features and global features separately, mapping them to a hidden dimension \(d\):
    \begin{equation}
        \mathbf{H}_{\text{node}} = \mathcal{X} \mathbf{W}_{\text{node}} + \mathbf{b}_{\text{node}} \in \mathbb{R}^{B \times T \times N \times d},
    \label{eq:linear_node}
    \end{equation}
    \begin{equation}
        \mathbf{H}_{\text{global}} = \mathcal{G} \mathbf{W}_{\text{global}} + \mathbf{b}_{\text{global}} \in \mathbb{R}^{B \times T \times d}.
    \label{eq:linear_global}
    \end{equation}
    
    \item \textbf{Global feature expansion and concatenation}: Replicate \(\mathbf{H}_{\text{global}}\) along the node dimension \(N\) times to obtain \(\mathbf{H}_{\text{global}}^{\text{exp}} \in \mathbb{R}^{B \times T \times N \times d}\), then concatenate it with the node features along the feature dimension:
    \begin{equation}
        \mathbf{H} = [\mathbf{H}_{\text{node}} \parallel \mathbf{H}_{\text{global}}^{\text{exp}}] \in \mathbb{R}^{B \times T \times N \times 2d},
    \label{eq:concat}
    \end{equation}
    where \(\parallel\) denotes concatenation. For brevity, we denote \(C = 2d\) as the unified feature dimension.
    
    \item \textbf{Positional encoding}: To incorporate temporal and spatial order information, we introduce learnable temporal positional encodings \(\mathbf{P}_{\text{time}} \in \mathbb{R}^{1 \times T \times 1 \times C}\) and spatial positional encodings \(\mathbf{P}_{\text{space}} \in \mathbb{R}^{1 \times 1 \times N \times C}\). These are added via broadcasting:
    \begin{equation}
        \mathbf{H} = \mathbf{H} + \mathbf{P}_{\text{time}} + \mathbf{P}_{\text{space}}.
    \label{eq:pos_enc}
    \end{equation}
    The positional encoding parameters are initialized to zero and optimized jointly with the model.
    
    \item \textbf{Dimension rearrangement}: The tensor shape is kept as \(\mathbb{R}^{B \times T \times N \times C}\) to match the input format of subsequent modules.
\end{enumerate}

Thus, the raw input is encoded into a feature tensor \(\mathbf{H}\) containing spatio-temporal information, ready for the following modules.

\subsection{Graph Propagating and Temporal Convolution}

\subsubsection{Graph Propagating}

To explicitly leverage graph structural information, we introduce a graph propagation module. The graph structure can be given by an external adjacency matrix \(\mathbf{A}_{\text{ext}} \in \mathbb{R}^{N \times N}\) or learned adaptively. Adaptive graph learning uses learnable node embeddings \(\mathbf{E} \in \mathbb{R}^{N \times d_a}\) to compute similarities via inner product, followed by ReLU activation, division by a learnable temperature parameter \(\tau\), and Softmax normalization:

\begin{equation}
    \hat{\mathbf{A}} = \text{Softmax}\left( \frac{\text{ReLU}(\mathbf{E} \mathbf{E}^\top)}{\tau} \right).
\label{eq:adaptive_adj}
\end{equation}

Given the normalized adjacency matrix, we perform \(K\)-step graph propagation. For the feature tensor \(\mathbf{H} \in \mathbb{R}^{B \times T \times N \times C}\), propagation is applied along the node dimension:

\begin{equation}
    \mathbf{H}^{(k)} = \hat{\mathbf{A}}^k \mathbf{H}, \quad k = 0, 1, \dots, K,
\label{eq:graph_prop}
\end{equation}
where \(\hat{\mathbf{A}}^0 = \mathbf{I}\) and \(\hat{\mathbf{A}}^k\) denotes the adjacency matrix after \(k\) propagation steps (implemented via matrix multiplication). In practice, \(\hat{\mathbf{A}}^k \mathbf{H}\) is realized by repeatedly applying \(\hat{\mathbf{A}}\) to \(\mathbf{H}\) along the node dimension.

Higher-order features (\(k \geq 1\)) are fused using learnable weights \(\{\alpha_k\}_{k=1}^K\), normalized by Softmax:

\begin{equation}
    \mathbf{H}_{\text{prop}} = \sum_{k=1}^K \alpha_k \mathbf{H}^{(k)}.
\label{eq:weighted_prop}
\end{equation}

\subsubsection{Temporal Convolution Module}

To capture local temporal dependencies for each station, we introduce a lightweight temporal convolution module. This module applies 1D convolutions along the time dimension to extract short-term variation patterns, and employs residual connections to enhance training stability.

Given the feature tensor \(\mathbf{H} \in \mathbb{R}^{B \times T \times N \times C}\), the temporal convolution module first reshapes it to \((B N) \times C \times T\), enabling parallel processing of all stations' time series along the merged batch-station dimension. A 1D convolution (either standard or depthwise separable) is then applied:

\begin{equation}
    \mathbf{H}_{\text{conv}} = \text{Conv1D}_{k_t, \text{groups}=g}\left( \mathbf{H}_{\text{reshaped}} \right),
\label{eq:conv1d}
\end{equation}
where the kernel size \(k_t\) controls the local receptive field; when \(g = C\), this becomes a depthwise separable convolution. The convolution output is sequentially passed through batch normalization, GELU activation, and Dropout:

\begin{equation}
    \mathbf{H}_{\text{conv}} = \text{Dropout}\left( \text{GELU}\left( \text{BN}(\mathbf{H}_{\text{conv}}) \right) \right).
\label{eq:conv_block}
\end{equation}

This module can be stacked for multiple layers to progressively capture local temporal features at different scales. The final output tensor retains the full spatio-temporal structure and can be directly fed into the subsequent agent attention module.

\subsection{Spatio-Temporal Agent Attention}

In spatio-temporal forecasting tasks, traditional self-attention faces a trade-off between computational complexity and long-range modeling capability. For traffic forecasting, the input typically comprises \(T\) time steps and \(N\) spatial nodes. Directly applying global self-attention to all \(T \times N\) tokens incurs a complexity of \(\mathcal{O}((TN)^2 C)\), which is infeasible for large-scale graphs. Even with separated spatial attention (self-attention among \(N\) nodes at each time step) and temporal attention (self-attention along time for each node), the total complexity remains \(\mathcal{O}(T N^2 C + N T^2 C)\), which is still costly when \(N\) or \(T\) is large. To address this, we employ the agent attention mechanism~\cite{han2024agent} in the spatio-temporal setting. Specifically, we introduce separate sets of spatial and temporal agent tokens, and decompose the agent attention into two symmetric branches that respectively model cross-station and cross-time dependencies. This design achieves efficient global modeling with linear complexity while preserving the expressiveness of Softmax attention.

\subsubsection{General Self-Attention Module}

Existing Transformer-based spatio-temporal forecasting methods typically adopt two attention paradigms. One is joint spatio-temporal attention, which flattens the spatio-temporal positions into \(N \times T\) tokens and computes self-attention among all tokens. Here, \(\mathbf{Q}, \mathbf{K}, \mathbf{V}\) are obtained by linearly projecting \(\mathcal{X}\). The complexity of this operation is \(\mathcal{O}((NT)^2 d)\), and memory consumption grows sharply with the number of stations or the time window length.

The second paradigm is separated spatio-temporal attention, which performs spatial and temporal attention independently. For spatial self-attention, the batch and time dimensions are merged into \((B \cdot T, N, C)\), yielding a complexity of \(\mathcal{O}(N^2 T d)\). Subsequently, temporal self-attention is applied independently for each node, resulting in \(\mathcal{O}(T^2 N d)\). The total complexity is \(\mathcal{O}(N^2 T d + T^2 N d)\).

Compared to the first approach, separated attention effectively models spatio-temporal dependencies separately, avoiding the high computational cost and memory footprint of joint attention. However, it still encounters a computational bottleneck when \(N\) or \(T\) is large.

\begin{equation}
\text{Attn}(\mathcal{X}) = \text{Softmax}\left(\frac{\mathbf{Q}\mathbf{K}^\top}{\sqrt{d_k}}\right)\mathbf{V},
\label{eq:standard_attn}
\end{equation}

\subsubsection{Basic Form of Spatio-Temporal Agent Attention}

For an input feature tensor \(X \in \mathbb{R}^{B \times T \times N \times C}\), we first view it as \(B \times T\) independent spatial graphs, each containing \(N\) nodes. Linear projections yield queries \(Q\), keys \(K\), and values \(V \in \mathbb{R}^{B \times T \times N \times C}\). To reduce computation, we introduce a small set of learnable spatial agent tokens \(P_s \in \mathbb{R}^{1 \times n_s \times C}\) (with \(n_s \ll N\)), which are broadcast to match the batch dimension. The spatial agent attention is computed in two steps:

\begin{enumerate}[leftmargin=*]
    \item \textbf{Agent aggregation}: The agent tokens \(P_s\) serve as queries, performing Softmax attention with keys \(K\) and values \(V\) to aggregate global information into agent context \(V_P\):
    \begin{equation}
    V_P = \text{Softmax}\big( (P_s) K^T / \sqrt{d} + B_{\text{agent}} \big) V,
    \label{eq:agent_agg}
    \end{equation}
    where \(B_{\text{agent}} \in \mathbb{R}^{1 \times H \times n_s \times 1}\) is an agent bias (\(H\) is the number of attention heads) and \(d = C/H\) is the dimension per head.

    \item \textbf{Agent broadcasting}: The original queries \(Q\) attend to the agent tokens \(P_s\) (now serving as keys) and the agent context \(V_P\) in a second Softmax attention, broadcasting the aggregated global information back to each node:
    \begin{equation}
    O_s = \text{Softmax}\big( Q (P_s)^T / \sqrt{d} + B_{\text{node}} \big) V_P,
    \label{eq:agent_broadcast}
    \end{equation}
    where \(B_{\text{node}} \in \mathbb{R}^{1 \times H \times 1 \times n_s}\) is another set of agent biases.
\end{enumerate}

To preserve feature diversity, we introduce a depthwise convolution (DWC) module applied to the values \(V\), and add it as a residual branch to the output:
\begin{equation}
O_s = O_s + \operatorname{DWC}(V),
\label{eq:dwc_residual}
\end{equation}
where DWC employs a depthwise separable 1D convolution with kernel size 3 (\texttt{Conv1d}, \texttt{groups = C}) sliding along the node dimension, effectively recovering feature diversity that may be lost in the two-step attention.

Similarly, for the temporal dimension, we introduce \(n_t\) learnable temporal agent tokens \(P_t \in \mathbb{R}^{1 \times n_t \times C}\) and transpose the input tensor to \((B, N, T, C)\), yielding \(B \times N\) time series of length \(T\). The temporal agent attention process is symmetric to the spatial version, using independent temporal agent biases \(B_{t,\text{agent}}\) and \(B_{t,\text{node}}\) to obtain the temporal attention output \(O_t\). DWC is not used in the temporal branch because the feature dimension is typically shorter and the loss of feature diversity is less significant.

\subsubsection{Overall Module Structure}

In the GraphAgentAttention module, when the input is a 4D tensor \((B, T, N, C)\) and temporal attention is enabled, the forward process is:
\begin{enumerate}[leftmargin=*]
    \item Apply spatial agent attention to the input, yielding \(X_{\text{spatial}} = \text{SpatialAttn}(X)\).
    \item Apply temporal agent attention to \(X_{\text{spatial}}\), yielding \(X_{\text{temporal}} = \text{TemporalAttn}(X_{\text{spatial}})\).
    \item The final output is the residual sum: \(X_{\text{out}} = X_{\text{spatial}} + X_{\text{temporal}}\).
\end{enumerate}

This module can be stacked for multiple layers, with residual connections and layer normalization to stabilize training.

\subsubsection{Complexity Analysis}

In spatial agent attention, the two attention steps each have a complexity of \(\mathcal{O}(B T N n_s C)\). Including the projections and DWC, the total complexity is:
\begin{equation}
\Omega = \underbrace{4 B T N C^2}_{\text{projections}} + \underbrace{B T N C}_{\text{agent acquisition}} + \underbrace{2 B T N n_s C}_{\text{agent attention}} + \underbrace{k^2 B T N C}_{\text{DWC}},
\label{eq:complexity}
\end{equation}
where \(k=3\) is the DWC kernel size. Temporal agent attention has a similar complexity of \(\mathcal{O}(B N T n_t C)\). Since \(n_s\) and \(n_t\) are constants much smaller than \(N\) and \(T\) (e.g., \(n_s = 32\), \(n_t = 32\)), the overall complexity is linear in both \(N\) and \(T\), i.e., \(\mathcal{O}(N T)\). In contrast, standard spatio-temporal self-attention (global or separated) is at least \(\mathcal{O}(N^2 T + N T^2)\), which becomes prohibitively expensive for large \(N\) or \(T\).

\subsubsection{Comparison with Standard Spatio-Temporal Attention}

Standard spatio-temporal attention typically adopts two paradigms:
\begin{itemize}[leftmargin=*]
    \item \textbf{Global attention}: Flattens the spatio-temporal positions into \(NT\) tokens and computes fully-connected self-attention, with complexity \(\mathcal{O}((NT)^2 C)\). It maintains a full receptive field but is infeasible for large-scale data.
    \item \textbf{Separated attention}: First computes spatial attention independently for each time step (\(\mathcal{O}(T N^2 C)\)), then temporal attention independently for each node (\(\mathcal{O}(N T^2 C)\)). Although complexity is reduced to quadratic, it still involves \(N^2\) and \(T^2\) terms.
\end{itemize}

In contrast, spatio-temporal agent attention reduces the quadratic complexity to linear by introducing a small number of learnable agent tokens. Specifically, the spatial and temporal branches have complexities of \(\mathcal{O}(BTN n_s C)\) and \(\mathcal{O}(BNT n_t C)\), respectively, where \(n_s, n_t \ll N, T\) are constants. Thus, the overall complexity is linear in \(N\) and \(T\), i.e., \(\mathcal{O}(NT)\), significantly outperforming the quadratic complexity of standard attention. Moreover, the two-step Softmax mechanism preserves global information exchange; the agent biases allow flexible encoding of relative positional relationships between nodes and time steps; and the DWC module compensates for the inherent feature diversity loss in linear attention, striking a favorable balance between computational efficiency and expressiveness.

\section{Experiment}

In this section, we present the implementation details, dataset descriptions, evaluation metrics, and baseline methods for comparison. We then conduct experiments to assess the performance of STAGformer, including comparisons with baseline methods, ablation studies, and case studies under different scenarios.

\subsection{Data Description}

To evaluate our model, we collected and integrated data from two cities, combining bike-sharing operational records with meteorological data, points of interest (POIs), and road network information. These data collectively provide the necessary input features for model training, as summarized in Table~\ref{tab:poi}.

\noindent\textbf{Bike-sharing trip data:} This dataset contains operational records from the NYC Citi-Bike\footnote{\url{https://ride.citibikenyc.com/system-data}} and Chicago Divvy-Bike\footnote{\url{https://divvybikes.com/system-data}} systems. The dataset primarily consists of two parts: (i) rental and return trip records for each station, and (ii) station-level details such as spatial and temporal information.

\noindent\textbf{Meteorological feature data:} Weather data for both cities were collected from the NOAA National Weather Service database\footnote{\url{https://www.noaa.gov/}}, providing historical daily weather records. Each record describes basic weather conditions and categorizes daily weather into four main types: sunny, cloudy, rainy/snowy, and extreme weather.

\noindent\textbf{Road network and POI data:} POI data were obtained from OpenStreetMap (OSM)\footnote{\url{https://www.openstreetmap.org/}}, offering detailed information around stations. Table~\ref{tab:poi} lists the main POI types and their counts for each city. We adopted POIs within a 150-meter radius of each station as input features for the model. Additionally, road network and city boundary data were collected from the open data platforms of New York City\footnote{\url{https://opendata.cityofnewyork.us/}} and Chicago\footnote{\url{https://data.cityofchicago.org/}} for supplementary analysis.

\begin{table}[htbp]
    \centering
    \footnotesize
    \caption{POIs Data from OpenStreetMap}
    \label{tab:poi}
    \begin{tabular}{lrr}
        \toprule
        POI type & New York & Chicago \\
        \midrule
        parking space   & 91669 & 5822  \\
        platform        & 11574 & 13240 \\
        bench           & 18325 & 2338  \\
        parking         & 5841  & 13399 \\
        garden          & 8571  & 4298  \\
        stop position   & 6366  & 5438  \\
        bicycle parking & 9196  & 2459  \\
        restaurant      & 7751  & 2829  \\
        pitch           & 6146  & 3995  \\
        fast food       & 4556  & 2139  \\
        park            & 2068  & 1589  \\
        playground      & 1825  & 1657  \\
        school          & 1809  & 1494  \\
        \dots           & \dots & \dots \\
        \bottomrule
    \end{tabular}
\end{table}

\subsection{Implementation}

\noindent\textbf{Preparation.} We preprocessed the bike-sharing trip data for New York City and Chicago, including determining station spatio-temporal information, data cleaning, missing value imputation, outlier detection, and normalization. Weather data were aligned with trip data at an hourly temporal resolution. Road network and POI data for each city were merged to form an external adjacency matrix for the graph propagation module. Finally, all data were consolidated into a unified dataset.

\noindent\textbf{Device.} Experiments were conducted on the AutoDL platform, with an Intel(R) Xeon(R) Platinum 8260 CPU @ 2.40GHz, an RTX 4080 SUPER GPU (32GB VRAM), and 60GB of RAM. The software environment included PyTorch 2.8.0, Python 3.12, CUDA 12.8, and Ubuntu 22.04. The model was trained using the AdamW optimizer with an initial learning rate of \(1\times10^{-3}\), batch size of 8, for 300 epochs, employing early stopping to prevent overfitting.

\subsection{Evaluation Metrics}

\noindent\textbf{Accuracy metrics.} We evaluate model performance using Root Mean Square Error (RMSE) and Mean Absolute Error (MAE). These metrics collectively provide a robust framework to assess the model's capability to capture subtle variations in the data.
\begin{equation}
    \text{RMSE} = \sqrt{\frac {1}{n}\sum _{i=1}^{n}(Y_{i} - \hat {Y_{i}})^{2}}
\label{eq:rmse}
\end{equation}
\begin{equation}
    \text{MAE} = \frac {1}{n}\sum _{i=1}^{n}|Y_{i} - \hat {Y_{i}}|.
\label{eq:mae}
\end{equation}
where \(\hat{y}_i\) is the predicted value, and \(y_i\) is the ground truth. \(N\) denotes the number of stations. The ground truth for station \(i\) is denoted as \(y_i = [y_i^{\text{rented}}, y_i^{\text{returned}}]\), and the predicted value is \(\hat{y}_i = [\hat{y}_i^{\text{rented}}, \hat{y}_i^{\text{returned}}]\).

\noindent\textbf{Computational efficiency metrics.} In addition to prediction accuracy, we evaluate computational efficiency using two widely adopted metrics: the number of parameters (Params) and the number of floating-point operations (FLOPs). Params reflect model complexity and storage requirements, while FLOPs measure the computational cost during inference. We report the total multiply-accumulate operations (MACs) as computed by the \texttt{thop} library, where 1 MAC is considered as one FLOP in our context. For a fair comparison, all models are evaluated under the same input shape \((B, T, N, F_n)\) and on the same hardware. These metrics demonstrate the scalability and practical deployability of the model, especially in large-scale micro-mobility networks.

\begin{table}[!t]
    \centering
    \caption{Performance of each model on two city Sharing-bike Datasets}
    \label{tab:baseline_monthly}
    \footnotesize
    \begin{tabular}{l ccccc ccccc}
        \toprule
        & \multicolumn{4}{c}{NYC Citi-Bike} & \multicolumn{4}{c}{Chicago Divvy-Bike} \\
        \cmidrule(lr){2-5} \cmidrule(lr){6-9}
        Model & \multicolumn{2}{c}{Sept. 2025} & \multicolumn{2}{c}{Oct. 2025}  & \multicolumn{2}{c}{Sept. 2025} & \multicolumn{2}{c}{Oct. 2025}\\
        \cmidrule(lr){2-3} \cmidrule(lr){4-5} \cmidrule(lr){6-7} \cmidrule(lr){8-9}
        & RMSE & MAE & RMSE & MAE & RMSE & MAE & RMSE & MAE \\
        \midrule
        Linear Regression   & 0.6336 & 0.3673 & 0.6152 & 0.3715  & 1.2282 & 0.8595 & 0.6262 & 0.3309  \\
        Spatial Regression  & 1.1176 & 0.6256 & 0.9377 & 0.5546  & 1.1327 & 0.4163 & 0.8353 & 0.3620  \\
        GRU           & 0.5538 & 0.2996 & 0.5572 & 0.2872  & 0.7823 & 0.2303 & 0.6394 & 0.2636  \\
        Transformer           & 0.5524 & 0.2905 & 0.5426  & 0.2984 & 0.8267 & 0.2480 & 0.6784 & 0.2399  \\
        GAT           & 0.5681 & 0.3272 & 0.5853 & 0.3426 & 0.7849 & \underline{0.2236} & 0.7419 & 0.3064 \\
        BikeMAN     & 0.6429 & 0.3198 & 0.5832 & 0.3030 & 0.8411 & 0.2699 & \underline{0.6034} & \underline{0.2345}  \\
        STAEformer           & \underline{0.5106} & \underline{0.2660} & \underline{0.5145} & 0.2628 & 0.8678 & 0.2666 & 0.7603 & 0.2764 \\
        T-STAR       & 1.1373 & 0.6616 & 0.9497 & 0.5718 & 1.1502 & 0.4570 & 0.8445 & 0.4004 \\
        BGM       & 0.6670 & 0.3452 & 0.6208 & \underline{0.2579} & \underline{0.7680} & 0.2557 & 0.6496 & 0.2542 \\
        \midrule
        \textbf{STAGformer} & \textbf{0.4874} & \textbf{0.2375} & \textbf{0.4782} & \textbf{0.2532} & \textbf{0.7419} & \textbf{0.2464} & \textbf{0.5440} & \textbf{0.1911} \\
        \bottomrule
    \end{tabular}
    \vspace{4pt}
    
    \footnotesize Underlined values indicate the best results among compared models; bold values represent our model.
\end{table}

\subsection{Comparison to Baselines}

We selected nine representative methods for comparison in this study, described as follows:

\begin{itemize}[leftmargin=*]
    \item \textbf{Linear Regression} \cite{singhvi2015predicting}: Predicts bike demand using weather and spatial features with neighborhood-level aggregation, serving as a simple yet interpretable baseline.
    
    \item \textbf{Spatial Regression} \cite{faghih2016incorporating}: Models spatial lag and error structures to capture spatio-temporal dependencies in bike-sharing demand, accounting for interactions among neighboring stations.
    
    \item \textbf{GRU} \cite{cho2014learning}: A gated recurrent neural network that captures temporal dependencies in sequential data for each station independently, widely used for time series forecasting.
    
    \item \textbf{Transformer} \cite{vaswani2017attention}: A self-attention based architecture that models long-range dependencies across all stations and time steps jointly, offering strong expressiveness but high computational cost.
    
    \item \textbf{GAT} \cite{velivckovic2017graph}: Applies attention over graph neighborhoods to aggregate spatial information with learnable importance weights, enhancing the modeling of dynamic spatial relationships.
    
    \item \textbf{BikeMAN} \cite{yang2025micromobility}: A multi-level spatio-temporal attention neural network for station-level bike traffic prediction. It employs an encoder-decoder architecture with spatial attention on input features and temporal attention on hidden states, integrating weather and POI data to capture complex correlations.
    
    \item \textbf{STAEformer} \cite{liu2023spatio}: A Transformer variant with spatio-temporal adaptive embeddings that enhances traffic forecasting performance by learning node-specific temporal patterns without complex architectural modifications.
    
    \item \textbf{T-STAR} \cite{cheng2026t}: A context-aware Transformer framework for short-term probabilistic demand forecasting in dock-based shared micro-mobility systems, integrating external factors and uncertainty estimation.
    
    \item \textbf{BGM} \cite{zhao2025bgm}: A dynamic graph modeling approach for demand prediction in expanding bike-sharing networks, adapting to new stations and evolving mobility patterns through graph structure learning.
\end{itemize}

\subsection{Ablation Study}

To thoroughly validate the effectiveness of each core module in STAGformer, we conducted a series of ablation experiments on both the New York and Chicago datasets, examining the change in model performance after removing specific modules. The ablation variants are defined as follows:

\begin{itemize}[leftmargin=*]
    \item \textbf{w/o E (without External Features):} Removes global external features (weather, time, POIs, etc.), using only historical station trip data as input.
    \item \textbf{w/o G (without Graph Propagation):} Removes the graph propagation module, i.e., no explicit aggregation of spatial neighbor information using the graph structure, retaining only temporal convolution and agent attention.
    \item \textbf{w/o TC (without Time Convolution):} Removes the temporal convolution module, relying solely on graph propagation and agent attention to capture temporal dependencies.
    \item \textbf{w/o Attn (without Agent Attention):} Removes the agent attention module, retaining only graph propagation and temporal convolution; thus, the model cannot model global long-range spatio-temporal interactions.
    \item \textbf{STAGformer:} The full model incorporating all modules.
\end{itemize}

\begin{figure}[!htbp]
    \centering
    \includegraphics[width=\textwidth]{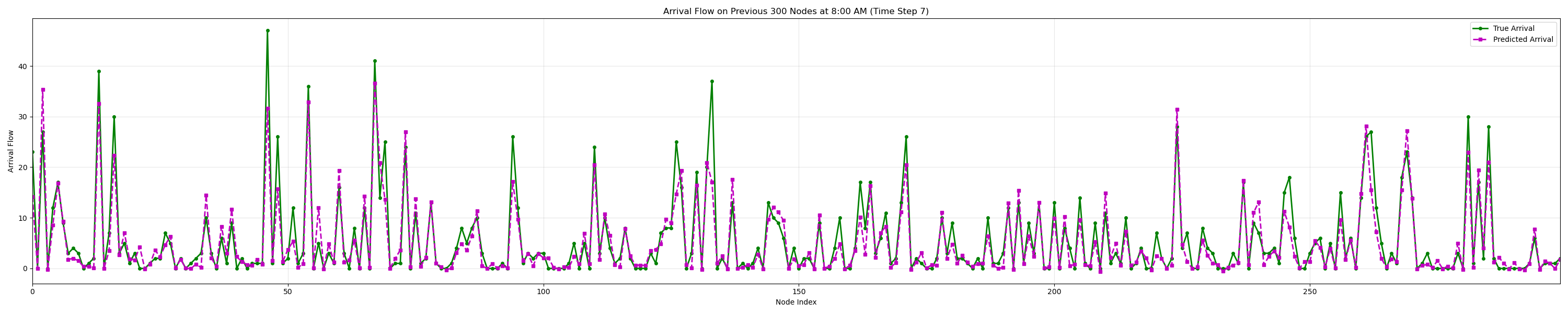}
    \includegraphics[width=\textwidth]{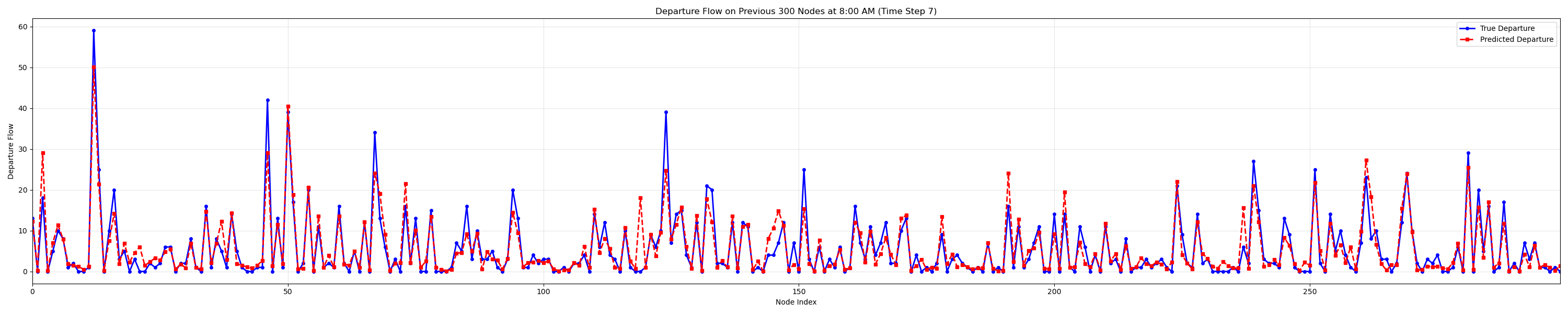}
    \caption{Arrival and departure patterns of bike-sharing stations in New York City.}
    \label{fig:arrive}
\end{figure}

\begin{figure}[htbp]
    \centering
    \includegraphics[width=0.85\textwidth]{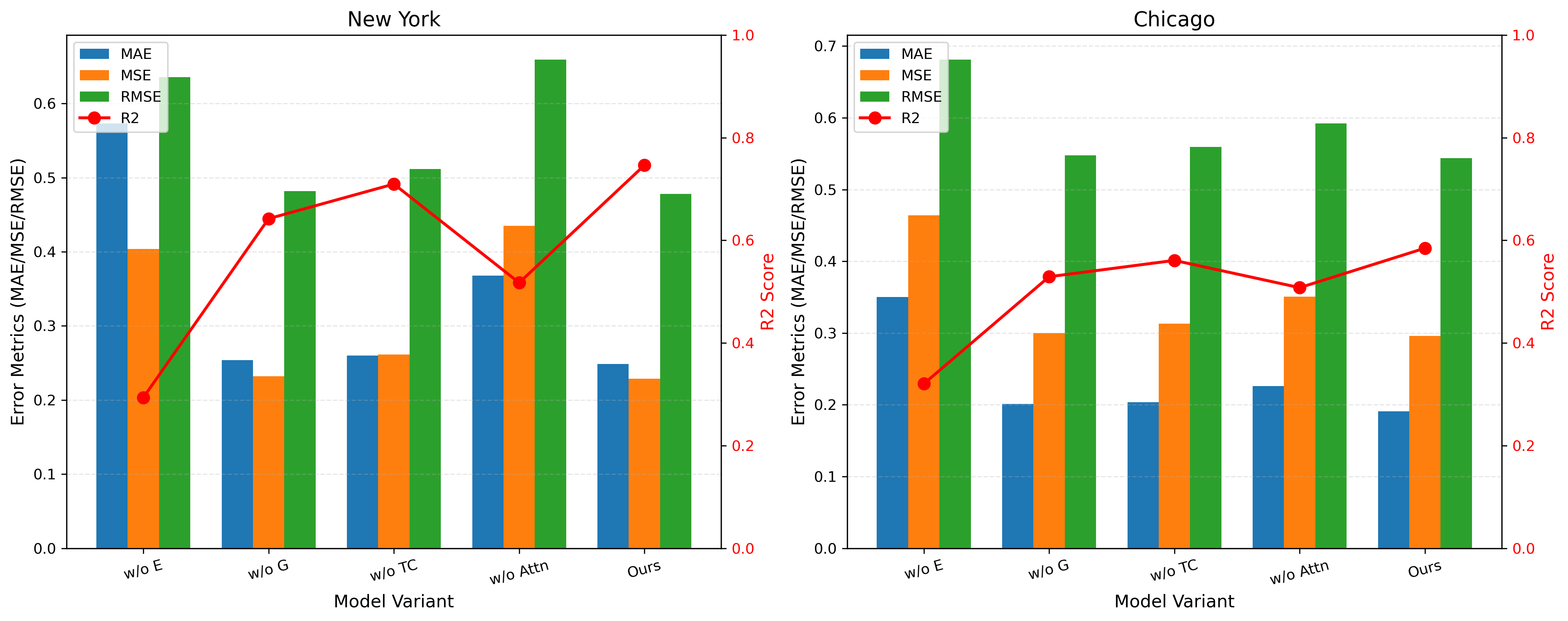}
    \caption{Ablation study results on New York and Chicago datasets. Bars represent error metrics (MAE, MSE, RMSE) and the red line shows the R\textsuperscript{2} score.}
    \label{fig:ablation}
\end{figure}

The experimental results, as shown in the figures, lead to the following conclusions:

\begin{enumerate}[leftmargin=*]
    \item \textbf{Importance of external features:} In the New York dataset, w/o External increased RMSE/MAE by 32.9\% / 130.7\% compared to the full model; in Chicago, the increases were 25.2\% / 83.2\%. This indicates that external factors such as weather, time, and POIs significantly influence bike-sharing demand, and fusing multi-source features effectively improves prediction accuracy.
    
    \item \textbf{Role of graph propagation module:} w/o Graph led to performance degradation in both datasets (New York: 0.8\% / 2.2\%; Chicago: 0.7\% / 5.3\%). This demonstrates that explicitly aggregating spatial neighbor information using the graph structure helps capture local spatial correlations, especially in areas with dense road network topology.
    
    \item \textbf{Necessity of temporal convolution module:} The error increases for w/o TConv (New York: 7.0\% / 4.6\%; Chicago: 2.8\% / 6.6\%) confirm the effectiveness of lightweight temporal convolution in extracting short-term temporal patterns for each station. Relying solely on attention mechanisms may overlook local details.
    
    \item \textbf{Contribution of agent attention module:} The most significant performance drop occurred for w/o Agent (New York: 37.9\% / 47.9\%; Chicago: 8.9\% / 18.4\%). This validates the critical role of agent attention in modeling global spatio-temporal dependencies, such as long-distance station coordination and periodic patterns. Removing this module reduces the model to only local graph propagation and temporal convolution, making it unable to capture complex global interactions.
\end{enumerate}

In summary, the four modules of STAGformer are complementary and indispensable. Global external features, graph propagation, temporal convolution, and agent attention characterize the spatio-temporal properties of bike-sharing demand from different dimensions, collectively ensuring high accuracy and robustness.

\subsection{Case Study}

\begin{figure}[!htbp]
    \centering
    \begin{subfigure}[b]{0.48\textwidth}
        \centering
        \includegraphics[width=\textwidth]{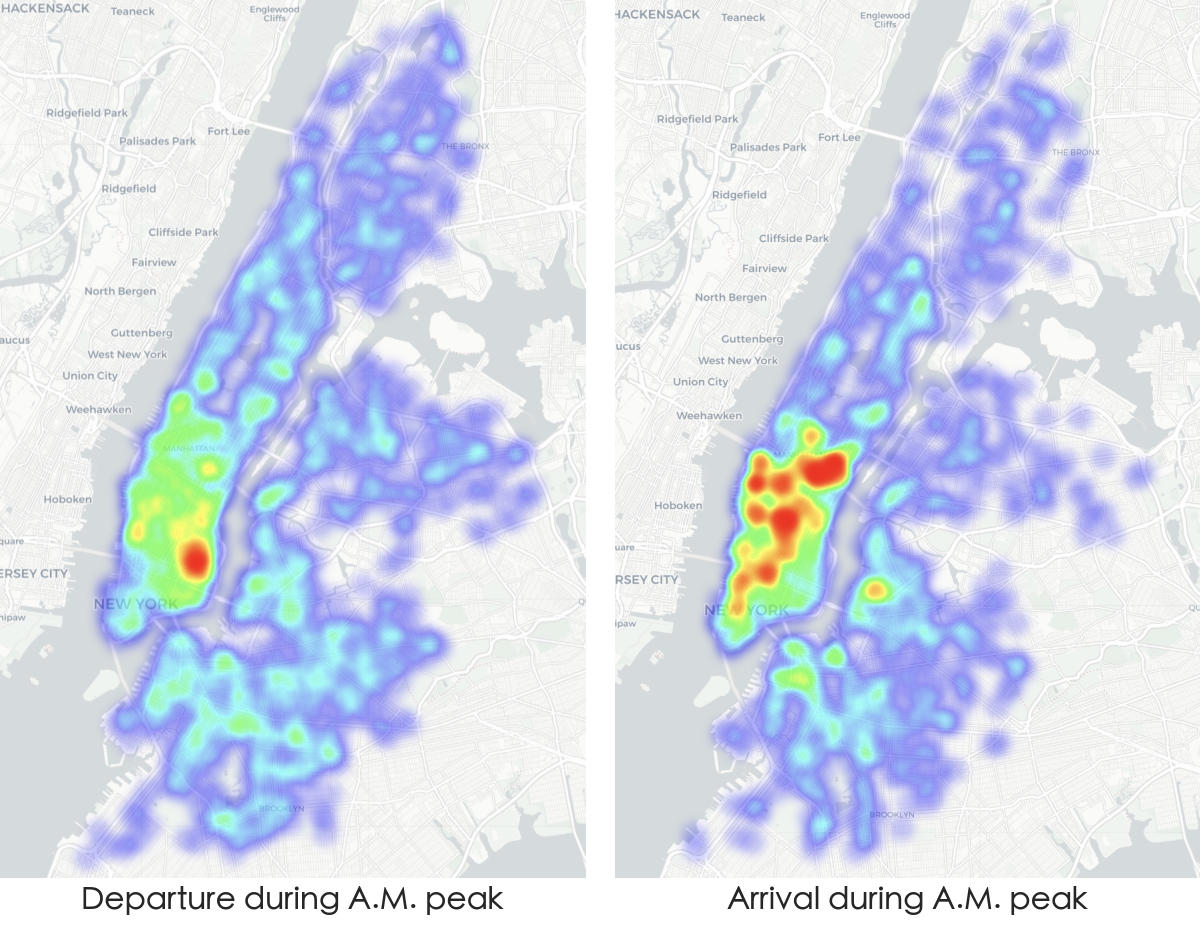}
        \caption{Morning peak}
    \end{subfigure}
    \hfill
    \begin{subfigure}[b]{0.48\textwidth}
        \centering
        \includegraphics[width=\textwidth]{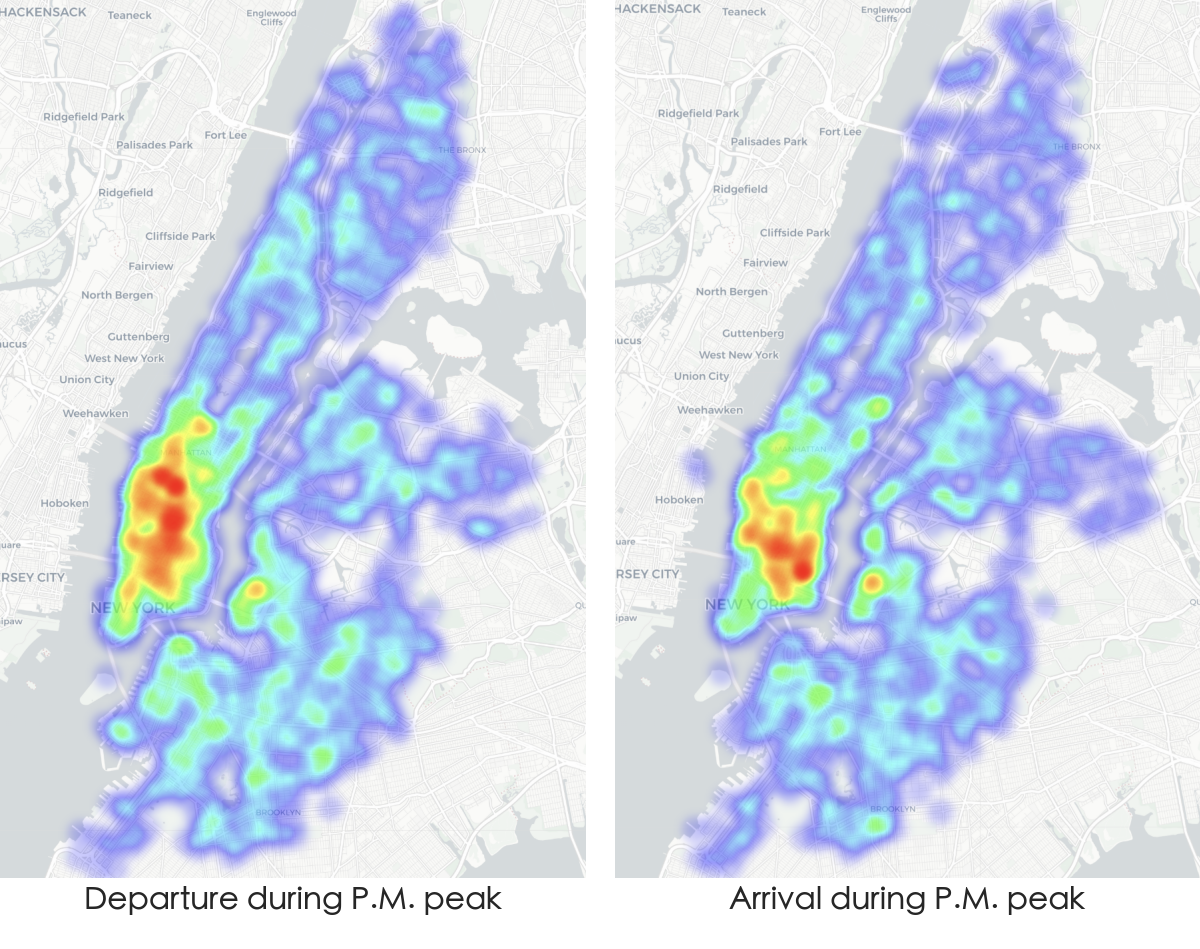}
        \caption{Evening peak}
    \end{subfigure}
    \caption{Forecast morning peak and evening peak departure and arrival heatmaps in New York City.}
    \label{fig:nyc_morning}
\end{figure}

In addition to the Chicago case presented in the introduction, we further validate the visualization of STAGformer on the New York City dataset. To intuitively illustrate the spatio-temporal heterogeneity of bike-sharing systems and verify STAGformer's ability to capture complex patterns, we plotted the departure and arrival distributions at 8:00 AM on a weekday in September 2025 for the NYC Citi-Bike system, as shown in Fig.~\ref{fig:arrive}. Since displaying all approximately 2000 stations simultaneously would make the image too dense to interpret, we randomly sampled 300 stations and compared the model's predictions with the actual trip demands. The upper subplot shows arrival distribution, and the lower subplot shows departure distribution. We also forecast the morning peak and evening peak, as shown in Fig.~\ref{fig:nyc_morning}; the color depth represents the magnitude of rentals/returns, with red indicating high demand and blue indicating low demand. The following typical patterns can be observed:

\begin{enumerate}[leftmargin=*]
    \item \textbf{High-intensity activity in core areas:} Stations in Midtown and Lower Manhattan exhibit extremely high departure and arrival volumes, reflecting commuter demand in the central business district during morning and evening peaks. These stations show a surge in departures during the morning peak and a surge in arrivals during the evening peak, exhibiting a pronounced tidal pattern.
    
    \item \textbf{Spatial imbalance:} In stark contrast to Manhattan, stations in Brooklyn and Queens (scattered points around Manhattan) have overall lower activity levels (mostly blue or light red), with relatively balanced departure and arrival distributions. This spatial heterogeneity stems from functional zoning: residential areas are dominated by departures in the morning peak, while commercial areas are dominated by arrivals; internal flows in peripheral residential areas are minimal.
    
    \item \textbf{Local coordination patterns:} Within Manhattan, neighboring stations often exhibit similar rental/return patterns. For example, stations near subway hubs maintain high turnover throughout the day, while stations in surrounding office areas show more pronounced morning and evening peaks. This local spatial correlation underscores the necessity of the graph propagation module.
    
    \item \textbf{Long-range dependencies between distant stations:} Although Manhattan and Brooklyn are geographically distant, stations at opposite ends of certain commuter corridors (e.g., Manhattan business district and Brooklyn residential areas) exhibit complementary departure and arrival patterns: high arrivals in Manhattan and high departures in Brooklyn during the morning peak, and the reverse during the evening peak. Such long-range dependencies exceed the receptive field of local graph convolution and require a global attention mechanism.
\end{enumerate}

STAGformer effectively captures the above patterns through the following design elements:
\begin{itemize}[leftmargin=*]
    \item \textbf{Graph propagation module} aggregates spatial neighbor information, strengthening local correlation modeling.
    \item \textbf{Agent attention module} models dependencies between distant stations at linear cost through spatial and temporal agents.
    \item \textbf{External feature fusion} encodes weather, time, and POI information, helping the model distinguish demand patterns in different functional zones.
\end{itemize}

The experimental results show that STAGformer achieves significantly lower prediction errors than baseline models in highly dynamic areas, demonstrating its capability to model complex spatio-temporal patterns. This case study not only visualizes the intrinsic structure of the data but also provides intuitive evidence for the effectiveness of the model design.

\section{Conclusion}

This paper proposed STAGformer, a spatio-temporal forecasting architecture that integrates the agent attention mechanism~\cite{han2024agent} with graph propagation and temporal convolution for station-level bike-sharing demand forecasting. By introducing separate spatial and temporal agent tokens and a two-step agent attention mechanism, the model reduces the quadratic complexity of standard self-attention to linear, while effectively capturing long-range dependencies across both spatial and temporal dimensions. The architecture integrates four key components: a spatio-temporal feature encoder that fuses dynamic node features with external global factors, a graph propagation module for aggregating spatial neighbor information, a temporal convolution module for modeling local temporal patterns, and a dedicated agent attention module that models global interactions via spatial and temporal agents. Extensive experiments on two real-world datasets (NYC Citi-Bike and Chicago Divvy-Bike) demonstrate that STAGformer consistently outperforms a wide range of baseline methods, achieving superior accuracy in terms of RMSE and MAE across multiple prediction horizons. Notably, the agent attention mechanism yields linear complexity with respect to the number of stations and time steps, enabling efficient deployment in large-scale urban networks without sacrificing the expressive power of softmax attention.

Ablation studies further validate the contribution of each module. Removing external features leads to significant performance drops, confirming the importance of integrating multi-source contextual information. The graph propagation and temporal convolution modules are shown to be essential for capturing local spatial correlations and short-term temporal dynamics, respectively. Most importantly, the agent attention module proves critical for modeling global spatio-temporal dependencies; its removal causes the largest error increase, underscoring its role in enabling long-range interactions among distant stations and capturing periodic patterns. Together, these components form a robust and efficient framework for micro-mobility demand forecasting.

Despite its promising results, STAGformer has certain limitations that open avenues for future research. First, the current model relies on a fixed graph structure derived from road network distances; incorporating dynamic graph learning that adapts to evolving mobility patterns could further improve accuracy. Second, while the agent attention mechanism reduces computational cost, its performance may be sensitive to the number of agent tokens, which requires empirical tuning. Future work could explore adaptive strategies for agent selection. Additionally, extending the model to other micro-mobility systems such as e-scooters or ride-hailing services, and incorporating richer external data (e.g., real-time events, social media trends), would test its generalizability. Finally, integrating STAGformer into a real-time decision-support system for bike-sharing operators could translate predictive gains into tangible operational benefits, such as optimized rebalancing and improved user satisfaction.

\section*{Acknowledgement}
This research is not supported by any organizations.




    


\begin{thebibliography}{99}

\bibitem{kim2025comprehensive}
J. Kim, H. Kim, H. Kim, D. Lee, and S. Yoon, ``A comprehensive survey of deep learning for time series forecasting: architectural diversity and open challenges,'' \textit{Artif. Intell. Rev.}, vol.~58, no.~7, p.~216, 2025.

\bibitem{chatfield2019analysis}
C. Chatfield and H. Xing, \textit{The Analysis of Time Series: An Introduction with R}, 7th ed. Boca Raton, FL, USA: Chapman and Hall/CRC, 2019.

\bibitem{lee2009advances}
D. D. Lee, P. Pham, Y. Largman, and A. Ng, ``Advances in neural information processing systems 22,'' \textit{Neural Inf. Process. Syst.}, vol.~1, no.~1, pp.~1--11, 2009.

\bibitem{zhou2021machine}
Z.-H. Zhou, \textit{Machine Learning}. Springer Nature, 2021.

\bibitem{friedman2001greedy}
J. H. Friedman, ``Greedy function approximation: a gradient boosting machine,'' \textit{Ann. Statist.}, pp.~1189--1232, 2001.

\bibitem{hochreiter1997long}
S. Hochreiter and J. Schmidhuber, ``Long short-term memory,'' \textit{Neural Comput.}, vol.~9, no.~8, pp.~1735--1780, 1997.

\bibitem{cho2014learning}
K. Cho et al., ``Learning phrase representations using RNN encoder--decoder for statistical machine translation,'' in \textit{Proc. Conf. Empir. Methods Nat. Lang. Process. (EMNLP)}, 2014, pp.~1724--1734.

\bibitem{zhang2017deep}
J. Zhang, Y. Zheng, and D. Qi, ``Deep spatio-temporal residual networks for citywide crowd flows prediction,'' in \textit{Proc. AAAI Conf. Artif. Intell.}, vol.~31, no.~1, 2017.

\bibitem{li2017diffusion}
Y. Li, R. Yu, C. Shahabi, and Y. Liu, ``Diffusion convolutional recurrent neural network: Data-driven traffic forecasting,'' \textit{arXiv preprint arXiv:1707.01926}, 2017.

\bibitem{yu2017spatio}
B. Yu, H. Yin, and Z. Zhu, ``Spatio-temporal graph convolutional networks: A deep learning framework for traffic forecasting,'' \textit{arXiv preprint arXiv:1709.04875}, 2017.

\bibitem{guo2019attention}
S. Guo, Y. Lin, N. Feng, C. Song, and H. Wan, ``Attention based spatial-temporal graph convolutional networks for traffic flow forecasting,'' in \textit{Proc. AAAI Conf. Artif. Intell.}, vol.~33, no.~1, 2019, pp.~922--929.

\bibitem{zheng2020gman}
C. Zheng, X. Fan, C. Wang, and J. Qi, ``GMAN: A graph multi-attention network for traffic prediction,'' in \textit{Proc. AAAI Conf. Artif. Intell.}, vol.~34, no.~1, 2020, pp.~1234--1241.

\bibitem{chen2021trafficstream}
X. Chen, J. Wang, and K. Xie, ``TrafficStream: A streaming traffic flow forecasting framework based on graph neural networks and continual learning,'' \textit{arXiv preprint arXiv:2106.06273}, 2021.

\bibitem{wu2019graph}
Z. Wu, S. Pan, G. Long, J. Jiang, and C. Zhang, ``Graph WaveNet for deep spatial-temporal graph modeling,'' \textit{arXiv preprint arXiv:1906.00121}, 2019.

\bibitem{bai2020adaptive}
L. Bai, L. Yao, C. Li, X. Wang, and C. Wang, ``Adaptive graph convolutional recurrent network for traffic forecasting,'' in \textit{Adv. Neural Inf. Process. Syst.}, vol.~33, 2020, pp.~17804--17815.

\bibitem{vaswani2017attention}
A. Vaswani et al., ``Attention is all you need,'' in \textit{Adv. Neural Inf. Process. Syst.}, vol.~30, 2017.

\bibitem{xu2020spatial}
M. Xu et al., ``Spatial-temporal transformer networks for traffic flow forecasting,'' \textit{arXiv preprint arXiv:2001.02908}, 2020.

\bibitem{child2019generating}
R. Child, S. Gray, A. Radford, and I. Sutskever, ``Generating long sequences with sparse transformers,'' \textit{arXiv preprint arXiv:1904.10509}, 2019.

\bibitem{katharopoulos2020transformers}
A. Katharopoulos, A. Vyas, N. Pappas, and F. Fleuret, ``Transformers are RNNs: Fast autoregressive transformers with linear attention,'' in \textit{Proc. Int. Conf. Mach. Learn. (ICML)}, 2020, pp.~5156--5165.

\bibitem{singhvi2015predicting}
D. Singhvi, S. Singhvi, P. I. Frazier, S. G. Henderson, E. O'Mahony, D. B. Shmoys, and D. B. Woodard, ``Predicting bike usage for New York City's bike sharing system,'' in \textit{Proc. AAAI Workshop on Computational Sustainability}, Austin, TX, USA, 2015.

\bibitem{choromanski2020rethinking}
K. Choromanski et al., ``Rethinking attention with performers,'' \textit{arXiv preprint arXiv:2009.14794}, 2020.

\bibitem{velivckovic2017graph}
P. Veli\v{c}kovi\'{c}, G. Cucurull, A. Casanova, A. Romero, P. Li\`{o}, and Y. Bengio, ``Graph attention networks,'' \textit{arXiv preprint arXiv:1710.10903}, 2017.

\bibitem{brody2021attentive}
S. Brody, U. Alon, and E. Yahav, ``How attentive are graph attention networks?'' \textit{arXiv preprint arXiv:2105.14491}, 2021.

\bibitem{zhang2018gaan}
J. Zhang, X. Shi, J. Xie, H. Ma, I. King, and D.-Y. Yeung, ``GaAN: Gated attention networks for learning on large and spatiotemporal graphs,'' \textit{arXiv preprint arXiv:1803.07294}, 2018.

\bibitem{wu2020connecting}
Z. Wu, S. Pan, G. Long, J. Jiang, X. Chang, and C. Zhang, ``Connecting the dots: Multivariate time series forecasting with graph neural networks,'' in \textit{Proc. ACM SIGKDD Int. Conf. Knowl. Discov. Data Min.}, 2020, pp.~753--763.

\bibitem{zhou2021informer}
H. Zhou et al., ``Informer: Beyond efficient transformer for long sequence time-series forecasting,'' in \textit{Proc. AAAI Conf. Artif. Intell.}, vol.~35, no.~12, 2021, pp.~11106--11115.

\bibitem{lee2019set}
J. Lee, Y. Lee, J. Kim, A. Kosiorek, S. Choi, and Y. W. Teh, ``Set transformer: A framework for attention-based permutation-invariant neural networks,'' in \textit{Proc. Int. Conf. Mach. Learn. (ICML)}, 2019, pp.~3744--3753.

\bibitem{behroozi2025predicting}
A. Behroozi and A. Edrisi, ``Predicting travel demand of a bike sharing system using graph convolutional neural networks,'' \textit{Public Transp.}, vol.~17, no.~1, pp.~281--317, 2025.

\bibitem{rochas2023contextual}
R. Rochas, A. Furno, and N.-E. El Faouzi, ``Contextual data integration for bike-sharing demand prediction with graph neural networks in degraded weather conditions,'' in \textit{Proc. IEEE Int. Conf. Intell. Transp. Syst. (ITSC)}, 2023, pp.~5436--5441.

\bibitem{wang2023demand}
J. Wang, T. Miwa, and T. Morikawa, ``A demand truncation and migration poisson model for real demand inference in free-floating bike-sharing system,'' \textit{IEEE Trans. Intell. Transp. Syst.}, vol.~24, no.~10, pp.~10525--10536, Oct.~2023.

\bibitem{han2024agent}
D. Han et al., ``Agent attention: On the integration of softmax and linear attention,'' in \textit{Proc. Eur. Conf. Comput. Vis. (ECCV)}, 2024, pp.~124--140.

\bibitem{feng2024adaptive}
J. Feng and H. Liu, ``An adaptive spatial-temporal method capturing for short-term bike-sharing prediction,'' \textit{IEEE Trans. Intell. Transp. Syst.}, vol.~25, no.~11, pp.~16761--16774, Nov.~2024.

\bibitem{zhao2025bgm}
Y. Zhao, H. Wen, X. Zhang, and M. Luo, ``BGM: Demand prediction for expanding bike-sharing systems with dynamic graph modeling,'' in \textit{Proc. Int. Joint Conf. Artif. Intell. (IJCAI)}, 2025, pp.~10008--10016.

\bibitem{xiang2025bike}
Z. Xiang, F. Zeng, L. Liu, J. Wu, S. Mumtaz, and V. C. M. Leung, ``Bike-sharing demand prediction based on dynamic time warping and spatio-temporal graph attention network,'' \textit{IEEE Trans. Intell. Transp. Syst.}, 2025, to be published.

\bibitem{jiang2023dual}
S. Jiang, Z. Strout, B. He, D. Peng, P. B. Shull, and B. P. L. Lo, ``Dual stream meta learning for road surface classification and riding event detection on shared bikes,'' \textit{IEEE Trans. Syst. Man Cybern. Syst.}, vol.~53, no.~11, pp.~7188--7200, Nov.~2023.

\bibitem{wiedemann2025geot}
N. Wiedemann, T. Uscidda, and M. Raubal, ``GeOT: a spatially explicit framework for evaluating spatio-temporal predictions,'' \textit{Int. J. Geogr. Inf. Sci.}, vol.~39, no.~10, pp.~2236--2266, 2025.

\bibitem{faghih2016incorporating}
A. Faghih-Imani and N. Eluru, ``Incorporating the impact of spatio-temporal interactions on bicycle sharing system demand: A case study of New York CitiBike system,'' \textit{J. Transp. Geogr.}, vol.~54, pp.~218--227, 2016.

\bibitem{yang2025micromobility}
X. Yang, J. Wang, S. Han, and S. He, ``Micromobility flow prediction: A bike sharing station-level study via multi-level spatial-temporal attention neural network,'' \textit{arXiv preprint arXiv:2507.16020}, 2025.

\bibitem{tziorvas2026mode}
A. Tziorvas, G. S. Theodoropoulos, and Y. Theodoridis, ``MoDE-Boost: Boosting shared mobility demand with edge-ready prediction models,'' \textit{arXiv preprint arXiv:2602.16573}, 2026.

\bibitem{liu2023spatio}
H. Liu et al., ``Spatio-temporal adaptive embedding makes vanilla transformer sota for traffic forecasting,'' in \textit{Proc. ACM Int. Conf. Inf. Knowl. Manag. (CIKM)}, 2023, pp.~4125--4129.

\bibitem{wang2410stgformer}
H. Wang, J. Chen, T. Pan, Z. Dong, L. Zhang, R. Jiang, and X. Song, ``STGformer: Efficient spatiotemporal graph transformer for traffic forecasting,'' \textit{arXiv preprint arXiv:2410.00385}, 2024.

\bibitem{liu2025stgformer}
Y. Liu and Z. Zhang, ``STGFormer: Spatio-temporal GraphFormer for 3D human pose estimation in video,'' \textit{Pattern Recognit.}, p.~112239, 2025.

\bibitem{cheng2026t}
J. Cheng, G. H. de Almeida Correia, O. Cats, and S. S. Azadeh, ``T-STAR: A context-aware transformer framework for short-term probabilistic demand forecasting in dock-based shared micro-mobility,'' \textit{arXiv preprint arXiv:2602.06866}, 2026.

\end{thebibliography}
\end{document}